\documentclass{article}

\usepackage{microtype}
\usepackage{graphicx}
\usepackage{subcaption}
\usepackage{booktabs} 

\usepackage{hyperref}

\usepackage[preprint]{icml2026}

\usepackage{amsmath}
\usepackage{amssymb}
\usepackage{mathtools}
\usepackage{amsthm}

\usepackage[capitalize,noabbrev]{cleveref}

\theoremstyle{plain}

\theoremstyle{definition}

\theoremstyle{remark}

\usepackage[textsize=tiny]{todonotes}

\usepackage[nolist]{acronym}
\usepackage{enumitem}
\usepackage{tcolorbox}
\tcbuselibrary{breakable}

\usepackage{listings}
\usepackage{xcolor}
\usepackage{makecell}
\usepackage{multirow}

\icmltitlerunning{Segment-Level Agentic Topic Modeling}

\begin{document}

\twocolumn[
  \icmltitle{Segment-Level Agentic Topic Modeling for Improved Data Exploration and Resource Efficiency}



  \icmlsetsymbol{equal}{*}

  \begin{icmlauthorlist}
    \icmlauthor{Myeongjun Erik Jang}{comp}
    \icmlauthor{Antonios Georgiadis}{comp}
    \icmlauthor{Sae Young Moon}{comp}
    \icmlauthor{Fran Silavong}{comp}

  \end{icmlauthorlist}

  \icmlaffiliation{comp}{J.P. Morgan Chase, London, United Kingdom}

  \icmlcorrespondingauthor{Myeongjun Erik Jang}{myeongjun.jang@jpmorganchase.com}
  \icmlcorrespondingauthor{Fran Silavong}{fran.silavong@jpmorganchase.com}

  \icmlkeywords{Machine Learning, ICML}

  \vskip 0.3in
]



\printAffiliationsAndNotice{}  

\begin{abstract}
    Topic modeling is an effective technique for discovering hidden themes within documents and is widely used in text mining and data analysis across a variety of industry sectors. Recently, large language model (LLM)-based topic models have been emerged that prompt LLMs to generate topics then assign the topics to documents, producing more natural and human-readable topics than conventional topic modeling algorithms. However, the nature of topic assignment process causes certain drawbacks, such as the incapability to produce topic distributions over a document, too broad or narrow topics, and high resource consumption, which increases with the number and length of of documents being assigned topics. These issues are particularly critical for industrial applications, which require high-quality, in-depth analysis and the processing of large volumes of documents. In this context, this paper introduces a framework called SeLATM, which addresses these concerns by employing segment-level topic generation and topic refinement through agentic feedback loops. Experimental results on various datasets demonstrate that SeLATM significantly reduces the LLM resources compared to methods based on topic assignment process, while maintaining superior performance.
\end{abstract}

\section{Introduction}
Topic modeling is a \ac{NLP} technique designed to discover meaningful topics within a corpus, which has a broad range of practical usage for data explorations in research and industry~\cite{ranaei2017topic, asmussen2019smart, xiong2019analyzing}.
Recent advances in \acp{LLM} have further propelled progress in this area. Compared to conventional methods such as LDA~\cite{LDA}, which represent a topic as distributions over words, these approaches prompt \acp{LLM} to generate free-text topics and assign the topic to individual documents~\cite{topicgpt, concept_induction, mu-etal-2024-large, doi-etal-2024-topic, LiSA, moon2026industryalignedgranulartopicmodeling}, resulting in more comprehensive and human-understandable topics.

Although, \ac{LLM}-driven topic generation and assignment methods provide significant benefits for the intuitiveness of topic representations, they are limited in their ability to capture document-topic relations. It is natural for a document to cover multiple topics. For example, a monthly financial report may address topics such as interest rates, global events, and political changes, with varying amounts of content devoted to each topic. Conventional topic modeling methods can capture this information, as they represent each document as a distribution over multiple topics. 
In contrast, \ac{LLM}-based approaches typically assign only a single topic to each document~\cite{concept_induction, LiSA, moon2026industryalignedgranulartopicmodeling}; even when multi-topic assignment is supported, these methods cannot precisely quantify how much each topic contributes to a document, limiting interpretability for data exploration. Also, the topic assignment process is highly resource-intensive, as it requires labeling topics to each individual document. Consequently, the \ac{LLM} call cost $O(N \times L)$ where $N$ and $L$ imply the number of documents and the document length. This poses a significant challenge for real-world industrial applications that process large volumes of data, particularly in sectors handling long documents, such as finance~\cite{masry2024longfin, hu2025finlbench} and legal~\cite{chalkidis-etal-2022-lexglue, jang-stikkel-2024-leveraging}.

Furthermore, current \ac{LLM}-based topic modeling approaches pay limited attention to topic refinement, either neglecting it at all~\cite{mu-etal-2024-large, doi-etal-2024-topic} or focusing solely on topic merging to reduce overlap~\cite{topicgpt, LiSA, moon2026industryalignedgranulartopicmodeling}. However, there is a potential risk that the generated topics maybe less coherent and could be further divided into smaller, more distinct topics. Additionally, relying solely on topic merging can result in ambiguous and vague topics. For example, consider two topics, $T_A$ and $T_B$, each comprising several subtopics ($t_i$): $T_A=\{t_i, 1 \leq i \leq 5\}$ and $T_A=\{t_i, 3 \leq i \leq 7\}$, respectively. Let us assume that the existing topic-merging methods combine these topic due to a significant overlap of 60\%. This results in a mega-topic $T=\{t_i, 1 \leq i \leq 7\}$, which could be too broad and incoherent, especially if subtopics $t_1$ and  $t_2$ are distinct from $t_6$ and $t_7$. Instead, it would be more natural to separate ${t_3, t_4, t_5}$ from $T_A$ and $T_B$, create a new topic, and then decide whether to merge it with the closer of $T_A$ or $T_B$; if not, the new topic should remain separate.

To this end, we propose a novel \ac{LLM}-driven agentic topic modeling algorithm called \textsc{SeLATM} (\textbf{Se}gment \textbf{L}evel \textbf{A}gentic \textbf{T}opic \textbf{M}odelling). As we regarded topic assignment as a major bottleneck of modern \ac{LLM}-driven topic modeling, we remove this step from our pipeline. Instead, to facilitate accurate topic naming from the outset, we introduce segment-level topic modeling, in which documents are divided into segments and topic names are assigned at the level of segment clusters. Owing to the semantic conciseness of text segments and their benefit to alleviate \acp{LLM}’ limitations when handling long contexts~\cite{liu-etal-2024-lost, zhou2024length}, our method can produce more precise topic names than previous approaches. In addition, our framework represents documents as distributions over multiple topics, allowing the contribution of each topic to a document and the association of specific content with corresponding topics, while also reducing computational costs. Furthermore, inspired by recent progress in iterative refinement via \ac{LLM}-driven feedback loops~\cite{madaan2023self, yuksel-etal-2025-multi, ravi2025llmloop}, we propose an agentic topic refinement process. In this framework, evaluator agents assess the quality of generated topics, operator agents apply topic split and merge operations based on the feedback from the evaluator, and a planner agent determines whether additional refinement iterations are required. To the best of our knowledge, this work is the first to explore the use of refinement-based feedback loops for topic modeling. Prior \ac{LLM}-based topic modeling approaches have been less amenable to such iterative refinement, as repeated topic assignment can substantially increase computational cost. Experiments on multiple public and industry datasets suggest that \textsc{SeLATM} yields higher-quality topics while remaining resource-efficient, thereby helping to mitigate the aforementioned concerns.

\section{\textsc{SeLATM} Framework}

\subsection{Segment-level Topic Modeling}

\textbf{Segmentation and Clustering.} Unlike a document that often spans multiple topics, a sentence or paragraph is typically structured to focus on a single core topic—a fundamental principle of written composition~\cite{strunk2007elements, moens2008using, halliday2013halliday}. This background motivates the following intuitions that underpin our approach:

\begin{enumerate}[nosep]
    \item As sentence- or paragraph-level segments will concentrate on a single topic, a cluster of semantically similar segments is expected to address the common topic underlying those segments.
    
    \item Assigning a label to a text cluster that conveys a single topic is considerably more straightforward than labeling a document that addresses multiple topics, whether using single or multiple labels.

    \item The reduced task complexity enables to achieve higher topic labeling accuracy, eliminating the need for document-level topic assignment process.

\end{enumerate}

Additionally, reduced token lengths offer several practical benefits: they help mitigate the lost-in-the-middle problem~\cite{liu-etal-2024-lost}, a well-known issue in long-context scenarios. Furthermore, it helps avoid length collapse~\cite{zhou2024length}—a phenomenon in which embeddings of long texts tend to cluster together and become less distinguishable—resulting in more precise clustering outcomes.


Let $D_i$ represent the $i$-th document in a corpus $\mathcal{D} = \{D_1, D_2, \ldots, D_N\}$. 
\textsc{SeLATM} first separates $D_i$ into text segments, $D_i = \{s_{i1}, s_{i2}, \ldots, s_{iM}\}$, where $s_{ij}$ denotes the $j$-th segment of $D_i$. Specifically, \textsc{SeLATM} adopts commonly used granularity for segmentation, sentence- and paragraph-level~\cite{DBLP:journals/corr/abs-2312-10997}, which can be generated using syntactic rules and do not require \ac{LLM} usage. We also experimented with recent semantic-based segmentation methods, such as \textsc{LumberChunker}~\cite{duarte-etal-2024-lumberchunker}; however, these approaches rely heavily on \acp{LLM}, which conflicts with our resource-efficiency objectives. Moreover, as sentences and paragraphs are typically organized around a single dominant topic, as mentioned above, we considered syntactic segmentation to be sufficient for our purposes. After segmentation, embeddings are generated for each text segment and subsequently clustered:
\begin{gather*}
e_{ij}=Emb(s_{ij}), \qquad \mathcal{C}=\{C_1, C_2, \ldots, C_{K}\} \\
f_c(e_{ij}) \mapsto C_k \in \mathcal{C}, \\
C_k = \{\, s_{ij} \mid f_c(e_{ij}) = k \,\}
\end{gather*}
where $e_{ij}$ denotes the embedding vector of $s_{ij}$ generated by the embedding model $Emb$, $C_k$ is $k$-th segment cluster, and $f_c$ is the clustering function that assigns $e_{ij}$ into its corresponding cluster. We used agglomerative clustering~\cite{murtagh2014ward}, as it produced a more stable and robust performance compared to K-means. The number of clusters ($K$) is a user-defined hyperparameter; increasing its value results in more granular topics. $K$ is treated as a granularity control rather than a ground-truth estimate, with downstream refinement mitigating moderate mis-specification.”

\textbf{Topic Generation. } The subsequent step involves generating topics for each text segment cluster. Regarding large size clusters, incorporating all text segments for topic generation is computationally inefficient. Therefore, to such clusters, \textsc{SeLATM} leverages Maximal Marginal Relevance (MMR, \citealt{10.1145/290941.291025}) sampling, which effectively extracts representative examples through an iterative selection process that balances the relevance and diversity of the selected documents~\cite{multiview_clustering}:
\begin{gather*}
\tilde{C}_k =
\begin{cases}
C_k & \text{if } |C_k| \leq n_S, \\
\mathrm{Sample}(C_k, n_S) & \text{otherwise},
\end{cases}
\end{gather*}
where $n_S$ is a hyperparameter that specifies the number of text segments sampled for topic generation. Finally, given the text segments of each cluster, a topic generation agent ($\mathcal{A}_{TG}$) generates topic names along with corresponding descriptions, thereby improving topic interpretability:
\begin{gather*}
(t_k, d_k) = \mathcal{A}_{TG}(\tilde{C}_k), \\
(t_k, d_k) \leftrightarrow C_k = \{\, s_{ij} \mid f_c(e_{ij}) = k \,\},
\end{gather*}
where $t_k$ and $d_k$ denote the topic name and description of cluster $k$. Sampling is applied exclusively before LLM calls and never before clustering or topic distribution computation. The prompt for topic generation is presented in Appendix~\ref{appendix.topic_gen_prompt}. 

As a result, \textsc{SeLATM} can represent a document $D_i$ as a distribution over topics as follows, where $p_{ik}$ is the contribution of $t_k$ to $D_i$:
\begin{gather*}
D_i = (p_{i1}, p_{i2}, \ldots , p_{iK}) \quad \text{where}, \\
p_{ik} = \frac{1}{|D_i|} \sum_{j=1}^{M} 1 (f_c(s_{ij})=C_k).
\end{gather*}
This enables us to capture the contribution of each topic and identify which segments are associated with certain topic-a capability not offered by previous \ac{LLM}-driven topic modeling methods. In addition, an \ac{LLM} is invoked $O(K \times n_S \times l)$ times, where $l$ is the length of the text segments. This is a significant reduction from $O(N \times L)$, given that $K \times n_S$ and $l$ are much smaller than $N$ and $L$, respectively.

\subsection{Agentic Topic Refinement}
Inspired by recent advances in leveraging \acp{LLM} to generate textual feedback~\cite{yuksekgonul2024textgrad, yin2025llm} and incorporate it into iterative refinement loops~\cite{madaan2023self, yuksel-etal-2025-multi, ravi2025llmloop}, we introduce an iterative topic refinement feedback loop that various agents collaborate. Although some applications demonstrate successful fully \ac{LLM}-driven process planning~\cite{song2023llm}, more comprehensive studies indicate that \acp{LLM} often perform poorly as planners~\cite{valmeekam2022large, valmeekam2023planning, kambhampati2024can, dagan2024dynamic}, and recommend manual curation of plans, roles, and prompts for a task requiring consistency~\cite{sypherd2024practical}. Hence, we designed a refinement process, which is presented in Algorithm~\ref{alg:agentic_refinement}.

\textbf{Topic Coherence Evaluation.} The first step is to measure topic coherence, which assess how closely the generated topic name and description are related to the text segments assigned to that topic. Low coherence indicates the present of text segments that do not align well with the assigned topic, suggesting that the topic may need to be separated. For each sampled segment cluster ($\tilde{C}_k$), the coherence evaluation agent ($\mathcal{E}_{coh}$) evaluates topic coherence in parallel using the cluster's segments, topic name, and description. If a topic is found to have low coherence, $\mathcal{E}_{coh}$ marks it as needing separation and provides feedback on how coherence can be improved. Detailed information regarding the prompt design of $\mathcal{E}_{coh}$ and its output format is provided in Appendix~\ref{appendix.topic_eval_coh}.

\begin{algorithm}[tb]
  \caption{Agentic Topic Refinement Feedback Loop}
  \label{alg:agentic_refinement}
  \begin{algorithmic}
    \STATE {\bfseries Input:} generated topics and descriptions $\mathcal{T}_g=\{(t_1, d_1), (t_2, d_2), \ldots, (t_K, d_K)\}$, segment clusters $\mathcal{C}=\{C_1, C_2, \ldots, C_K \}$, sampled segment clusters  $\mathcal{\tilde{C}}=\{\tilde{C_1}, \tilde{C_2}, \ldots, \tilde{C_K}\}$, number of iterations $e$, maximum tolerance $T$.

    \STATE{\bfseries Agents:} topic coherence evaluator agent $\mathcal{E}_{coh}$, topic diversity evaluator agent $\mathcal{E}_{div}$, split operation agent $\mathcal{A}_{split}$, merge operation agent $\mathcal{A}_{merge}$, planner agent $\mathcal{A}_{plan}$ 

    \STATE{\bfseries Output: } refined $\mathcal{C}$ and $\mathcal{T}_g$.
    
    \STATE \textbf{Initialize} $tolerance = 0$.
    \FOR{$i=1$ {\bfseries to} $e$}
    \IF{$tolerance > T$}
    \STATE break
    \ENDIF
    \STATE $\mathcal{C}^i, \mathcal{\tilde{C}}^i, \mathcal{T}_g^i =$ COPY$(\mathcal{C}, \mathcal{\tilde{C}}, \mathcal{T}_g)$
    
    \STATE \# Generate coherence evaluation feedback
    \STATE $Coh_{fb} = \mathcal{E}_{coh} (\mathcal{\tilde{C}}^i, \mathcal{T}_g^i)$
    \STATE \# Run split operation, and update $\mathcal{C}^i$, $\mathcal{\tilde{C}}^i$, and $\mathcal{T}_g^i$
    \STATE $\mathcal{C}^{i}, $ $\mathcal{T}_g^i$$ = \mathcal{A}_{split}(Coh_{fb}, \mathcal{C}^i, \mathcal{T}_g^i )$
    \STATE $\mathcal{\tilde{C}}^{i}$ = UPDATE($\mathcal{\tilde{C}}^i$, $\mathcal{C}^{i}$)

    \STATE \# Generate diversity evaluation feedback
    \STATE $Div_{fb} = \mathcal{E}_{div} (\mathcal{T}_g^i)$
    \STATE \# Run merge operation, and update $\mathcal{C}^i$, $\mathcal{\tilde{C}}^{i}$, and $\mathcal{T}_g^i$
    \STATE $\mathcal{\tilde{C}}^{i}, \mathcal{T}_g^i = \mathcal{A}_{merge}(Div_{fb}, \mathcal{\tilde{C}}^{i}, \mathcal{T}_g^i)$
    \STATE $\mathcal{C}^{i}$ = UPDATE($\mathcal{\tilde{C}}^i$, $\mathcal{C}^{i}$)

    \STATE \# Re-evaluate refined topics
    \STATE $Coh_{fb}^{r}, Div_{fb}^{r} = \mathcal{E}_{coh} (\mathcal{\tilde{C}}^i, \mathcal{T}_g^i), \mathcal{E}_{div} (\mathcal{T}_g^i)$ 

    \STATE $out = \mathcal{A}_{plan}(Coh_{fb}, Coh_{fb}^{r}, Div_{fb}, Div_{fb}^r)$
    \IF{$out.is\_improved$ is $true$}
    \STATE $\mathcal{C} = \mathcal{C}^i$ and $\mathcal{\tilde{C}} = \mathcal{\tilde{C}}^{i}$ and $\mathcal{T}_g = \mathcal{T}_g^i$
    \ELSE
    \STATE $tolerance$ += $1$    
    \ENDIF
    \ENDFOR
  \end{algorithmic}
\end{algorithm}

\textbf{Topic Split Operation. } For topics identified as needing separation, the split operation agent ($\mathcal{A}_{\text{split}}$) performs the separation. Specifically, the text segments of the topic ($C_k$) are re-clustered with the number of clusters $K$ set to 2. Please note that $C_k$ is used instead of $\tilde{C}_k$, because segments that are not included in the sampled segment cluster must be assigned to one of the new topics. Subsequently, we measure the proportion of text segments in the larger of the two separated clusters. If this proportion exceeds certain threshold (80\% in our experiments), we skip the split operation, as this indicates that, despite $\mathcal{A}_{\text{split}}$ recommending separation, the majority of text segments still pertain to a single topic. Otherwise, $\mathcal{A}_{TG}$ regenerates the topic names and descriptions for each separated segment cluster, this time incorporating the feedback provided by $\mathcal{E}_{coh}$ into the prompt. Once the names and descriptions are re-generated, we update $\mathcal{\tilde{C}}$ accordingly by removing the split topic and adding the new topics resulting from the separation.

\textbf{Topic Diversity Evaluation.} After the split operation, topic diversity is assessed by topic diversity evaluation agent ($\mathcal{E}_{div}$), which evaluates how distinctive each topic is from the others. This is a holistic, one-shot evaluation that takes all topic names and descriptions ($\mathcal{T}_g$) as input and identifies which topic groups should be merged, returning the corresponding topic indices along with feedback on how merging can improve topic diversity. Detailed information about the prompt used for $\mathcal{E}_{div}$ and its output format is provided in Appendix~\ref{appendix.topic_eval_div}.

\textbf{Topic Merging Operation.} Given that $\mathcal{E}_{div}$ identifies the need for topic merging and specifies the groups of topics to be united, the merge operation agent ($\mathcal{A}_{merge}$) combines the sampled text segments ($\tilde{C}_k$) of each identified topic group and run $\mathcal{A}_{TG}$ to generate the name and description of the merged topic. As with the topic split operation, feedback from $\mathcal{E}_{div}$ is incorporated into the topic generation prompt. Once the merging operation is completed for all identified topic groups, $\mathcal{C}$ is updated accordingly by replacing the merged topics with the newly formed topics.

\textbf{Planning Agent.} The refined topics are re-evaluated by $\mathcal{E}_{coh}$ and $\mathcal{E}_{div}$, and the results are then passed to the planning agent ($\mathcal{A}_{plan}$), which determines  whether improvements have been achieved. Specifically, the planning agent first invokes \ac{LLM} to generate the summary reports for both the pre- and post-refinement iteration. It then compares these reports to determine whether any improvement has been achieved. If an improvement is observed, the topic modeling results ($\mathcal{C}$, $\mathcal{\tilde{C}}$, and $\mathcal{T}_g$) are updated with the refined ones and the iteration continues. Otherwise, no updates are made and only the $tolerance$ parameter is incremented by 1. The refinement loop proceeds until either the maximum tolerance ($T$) or the maximum number of iterations ($e$) is reached.
Details regarding the prompts used for $\mathcal{A}_{plan}$ and its output format is provided in Appendix~\ref{appendix.planner_prompt}.

\subsection{Multi-view Topic Parenting} In addition to topics, \textsc{SeLATM} also provides parent topics because presenting higher-level topics along with their relationship with base topics can offer valuable insights from an analyst’s perspective. Unlike the topic generation stage, where only text segments are available, parent topics are established based on the topics themselves, allowing additional access to the topic names and descriptions. In this regard, we employ multi-view clustering method~\cite{multiview_clustering}. Specifically, the multi-view embedding of each topic ($e_k$) is constructed by concatenating the centroid of the text segment embeddings ($e_{c_k}$), the topic name embedding ($e_{t_k}$), and the topic description embedding ($e_{d_k}$). Additionally, the natural language representation of each topic is formed by concatenating the topic name and description. With these representations, we repeat agglomerative clustering to categorize similar topics and use $\mathcal{A}_{TG}$ to generate the name and description of parent topics.

\section{Experiment Design}

\begin{table}[t!]
  \caption{Basic statistics of datasets used for experiments.}
  \vspace{-2ex}
  
  \label{table.data_stats}
  \renewcommand{\arraystretch}{1.0}
  \footnotesize{
        \centering{\setlength\tabcolsep{1.0pt}}
    }
  \begin{center}
    \begin{small}
      \begin{sc}
        \begin{tabular}{cccc}
          \toprule
          Data set  & Data size   &  Token Len & \# of Labels \\
          \midrule
          Banking77 & 1,947 & 13 & 77 \\
          Bills     & 1,000 & 233 & 101 \\
          Wiki      & 1,100 & 3950 & 207 \\
          CCC       & 603 & 116 & - \\
          BCR       & 1,250 & 15 & - \\
          CI        & 3,000 & 17 & -  \\
          \bottomrule
        \end{tabular}
      \end{sc}
    \end{small}
  \end{center}
\vspace{-3ex}
\end{table}

\subsection{Datasets} 
We selected three publicly available datasets, Bills~\cite{bills} and Wiki~\cite{DBLP:journals/corr/abs-2312-10997}, which are widely used for topic modeling evaluation, and Banking77~\cite{banking77} for its specificity to the finance domain. As noted by \citealt{topicgpt}, using the entire training corpus for topic modeling is impractical. Therefore, we applied the same sampling strategy as \citealp{topicgpt} for our experiments. We also conduct experiments on three in-house business datasets. The \textbf{C}ustomer \textbf{C}hatbot \textbf{C}onversation (CCC) dataset consists of conversation messages from a consumer banking chatbot application. The \textbf{B}anking \textbf{C}hatbot \textbf{R}eview (BCR) dataset contains customer reviews of the baking chatbot application. The \textbf{C}ustomer \textbf{I}ssues (CI) dataset comprises short summaries describing issues raised by customers during their daily banking activities. All three industry datasets does not contain ground-truth labels. Table~\ref{table.data_stats} shows basic statistics of the datasets.

\subsection{Evaluation Metrics}
The public datasets include ground-truth labels, allowing the use of standard clustering-based evaluation metrics as employed in previous studies~\cite{bills, topicgpt}. Accordingly, we calculated the Harmonic Mean of Purity (P1), Adjusted Rand Index (ARI), and Normalized Mutual Information (NMI) metrics, with detailed explanations provided in Appendix~\ref{appendix:cluster_eval_metrics}. These metrics require a single prediction to be compared with the labels. Therefore, for \textsc{SeLATM}, we used the topic with the highest contribution as the prediction for evaluation. If there was a tie in top contributions, we treated it as a new topic.

Clustering-based metrics are limited in that they cannot be applied to datasets without ground-truth labels and cannot assess the quality of assigned topic names. Therefore, we evaluated topic accuracy ($\mathcal{TA}$) and topic completeness ($\mathcal{TC}$)~\cite{moon2026industryalignedgranulartopicmodeling}, which measure how accurately topic names are assigned to documents and whether any topics are missing from the documents, respectively. We employed \ac{LLM}-as-a-Judge building on its recent success on automated evaluations~\cite{NEURIPS2023_91f18a12, li-etal-2024-leveraging-large, jangsilavong2025instajudge}. As \ac{LLM} are prone to producing inconsistent predictions~\cite{jang-lukasiewicz-2023-consistency, wang2024-large-language-models-fair}, we used self-consistency decoding for more reliable evaluations~\cite{wangself}. Specifically, for each $\mathcal{TA}$ and $\mathcal{TC}$, we prompted \ac{LLM} to generate five predictions—each with a distinct reasoning path—given a document and its assigned topic name, following the label schema presented in Table~\ref{table.autoeval_label}. The predictions are converted to their corresponding scores, and the final score is computed as the average of these five scores. The prompts of $\mathcal{TA}$ and $\mathcal{TC}$ evaluations are presented in Appendix~\ref{appendix.llm_as_judge_prompt}. When the output consists of multiple topics, we concatenated the generated topics into a single text for measuring $\mathcal{TC}$, considering the definition of the metric. However, this strategy can provide additional benefits when measuring $\mathcal{TA}$. Since \textsc{SeLATM} provides a distribution of topics, we used the topic ratios as weights and calculated the weighted average of each individual topic's $\mathcal{TA}$ score. For methods that do not provide topic weights, we applied the same approach as used for $\mathcal{TC}$.

\begin{table}[t!]
  \caption{Textual labels and corresponding scores used for LLM-as-a-Judge evaluation.}
  \vspace{-2ex}
  
  \label{table.autoeval_label}
  \renewcommand{\arraystretch}{1.0}
  \footnotesize{
        \centering{\setlength\tabcolsep{2.0pt}}
    }
  \begin{center}
    \begin{small}
      \begin{sc}
        \begin{tabular}{ccc}
          \toprule
          $\mathcal{TA}$  & $\mathcal{TC}$ & Score \\
          \midrule
          Incorrect & Not Covered & 0 \\
          Partially Correct & Minorly Covered & 1/3 \\
          Mostly Correct & Mostly Covered & 2/3 \\
          Completely Correct & Complete & 1 \\
          \bottomrule
        \end{tabular}
      \end{sc}
    \end{small}
  \end{center}
\vspace{-4.5ex}
\end{table}

\subsection{Baseline Methods}
As baselines, we selected the following approaches from three distinct topic modeling families, based on their reported performance and code availability.

\begin{itemize}[nosep]
    \item\textbf{Embedded topic models}: BERTopic~\cite{bertopic}, CTop2Vec~\cite{contextual_top2vec}.
    \item \textbf{\ac{LLM}-driven topic generation and assignment}: TopicGPT~\cite{topicgpt}, LLoom~\cite{concept_induction}, TIDE~\cite{moon2026industryalignedgranulartopicmodeling}.
    \item \textbf{\ac{LLM}-assisted neural topic models}: LLM-ITL~\cite{LLM_ITL}.
\end{itemize}

Further details about the baselines and their implementations are provided in Appendix~\ref{appendix.baselines}.

\subsection{Models and Hyperparameters}
For experiments, we used OpenAI models; \texttt{gpt-4o-2024-08-06} for language generation and \texttt{text-embedding-3-small-1} for generating embedding vectors. The same \acp{LLM} were also applied to baselines unless their code implementations require specific models. Regarding \ac{LLM}-as-a-Judge, we used a different model, \texttt{gpt-4.1-mini-2025-04-14}.

The number of topics ($K$) is a necessary hyperparameter for \textsc{SeLATM}, TIDE, and LLM-ITL. Based on the number of ground-truth labels, we set $K$ to 100 for the Banking77 and Bills datasets, and to 250 for the Wiki dataset. For the business datasets, which do not contain ground-truth labels, we set $K$ to 50. \textsc{SeLATM} requires an additional hyperparameter, $n_S$, which specifies the number of sampled text segments for each cluster. For each dataset, we identified the best $n_S$ through hyperparameter search; detailed experimental results are provided in Appendix~\ref{appendix.hpo_Ns}.

\begin{table*}[t!]
    \caption{Experimental results on the public datasets. We report an average of 5 repetitions. The best performance for each evaluation metric is highlighted in bold; underlined values indicate statistically significant performance gap over the best-performing baselines (or over \textsc{SeLATM} when a baseline performs best), determined by t-test with a p-value $<$ 0.05. `\textsc{SeLATM} Iter:0' refers to \textsc{SeLATM} results without the refinement loop;  italicized values indicate where `\textsc{SeLATM} Iter:0' outperforms the baselines.}
    \label{table:pub_data_experiments}
    \vspace{-1ex}
    \begin{center}
        \renewcommand{\arraystretch}{1.0}
        \footnotesize{
            \centering{\setlength\tabcolsep{1pt}}
        }
        \begin{tabular}{cccccccccccccccc}
        \toprule
        \multirow{2}{*}{\textbf{Models}} & \multicolumn{5}{c}{\textbf{Banking77}} & \multicolumn{5}{c}{\textbf{Bills}} & \multicolumn{5}{c}{\textbf{Wiki}} \\
        & P1 & ARI & NMI & $\mathcal{TA}$ & $\mathcal{TC}$ & P1 & ARI & NMI & $\mathcal{TA}$ & $\mathcal{TC}$ & P1 & ARI & NMI & $\mathcal{TA}$ & $\mathcal{TC}$ \\ \midrule
        
        BERTopic & \underline{\textbf{.705}} & .468 & .817 & .634 & .667 & .359 & .230 & .558 & .099 & .066 & .303 & .144 & .652 & .137 & .076 \\

        C-Top2vec & .557 & .406 & .788 & .470 & .704 & .424 & .230 & \underline{\textbf{.767}} & .350 & .379 & .490 & .341 & .685 & .501 & .574 \\

        TopicGPT & .172 & .078 & .532 & .734 & .683 & .385 & .244 & .709 & .748 & .740 & .392 & .183 & .779 & .826 & \underline{\textbf{.627}} \\
        LLooM  & .111 & .037 & .322 & .283 & .259 & .159 & .048 & .295 & .501 & .424 & .110 & .046 & .424 & .544 & .353 \\ 

        TIDE & .624 & .486 & .800 & .782 & .867 & .440 & \underline{\textbf{.285}} & .711 & .759 & .719 & .389 & .259 & \textbf{.816} & .641 & .621 \\

        LLM-ITL  & .244 & .001 & .562 & .379 & .419 & .395 & .033 & .723 & .129 & .110 & .491 & .046 & .802 & .168 & .010 \\      
        \midrule

        \textsc{SeLATM} Iter:0 & .684 & \underline{\textit{\textbf{.542}}} & \underline{\textit{\textbf{.832}}} & \textit{.955} & \textit{.953} & \textit{.440} & .255 & .663 & \textit{.795} & \textit{.769} & \textit{.522} & \textit{.357} & .805 & \textit{.832} & .574 \\

        \textsc{SeLATM} Iter:1 & .680 & .539 & \underline{\textbf{.832}} & .962 & .956 & .440 & .260 & .661 & .803 & .782 & \underline{\textbf{.524}} & \underline{\textbf{.359}} & .803 & .807 & .567 \\

        \textsc{SeLATM} Iter:3 & .672 & .533 & .830 & \underline{\textbf{.966}} & \underline{\textbf{.960}} & \textbf{.443} & .258 & .660 & \underline{\textbf{.823}} & \underline{\textbf{.792}} & .522 & .355 & .803 & \underline{\textbf{.842}} & .587 \\

        \textsc{SeLATM} Iter:5 & .663 & .526 & .826 & .962 & .961 & .441 & .261 & .659 & .815 & .789 & .514 & .354 & .805 & .841 & .588 \\
        \bottomrule
        \end{tabular}
    \end{center}
    \vspace{-3ex}
\end{table*}

\section{Experimental Results}
In our experiments, we investigated different \textsc{SeLATM} variations by setting the number of iterations for the feedback refinement loop to 0, 1, 3, and 5.

\subsection{Public Dataset Results}
\textbf{Effect of Segment-level Topic Modeling.} The results of the public dataset experiments are presented in Table~\ref{table:pub_data_experiments}. The results show that \textsc{SeLATM} outperforms the baselines on the majority of evaluation metrics. Even without the feedback refinement loop, \textsc{SeLATM} achieves the best performance in 10 out of 15 evaluation cases, with 9 of these differences being statistically significant. The improvements are more striking in \ac{LLM}-as-a-judge metrics, especially for $\mathcal{TA}$, where \textsc{SeLATM} outperforms the best-performing baselines by an average of 21\%. This finding supports our intuition that generating topics for univocal text clusters reduces task complexity and leads to higher topic modeling accuracy. Moreover, in cases where the baseline models achieve the highest ranking, \textsc{SeLATM} mostly secures the second-best performance—for example, in P1 for Banking77, ARI for Bills, and NMI and $\mathcal{TC}$ for the Wiki dataset. Notably, \textsc{SeLATM} demonstrates a substantial margin over the third-ranked model in P1 for Banking77 and ARI for Bills. Although some baselines, such as TIDE and C-Top2vec, achieve the highest scores in certain evaluation cases, it is worth to mention that none of the baselines consistently outperform the others across majority of evaluation cases, as \textsc{SeLATM} does. The results highlight the advantage of segment-level topic modeling over the baselines, as it not only offers greater interpretability by providing multiple topics along with their contribution scores, but also outperforms the baselines across various evaluation metrics. Several output examples of the public datasets are provided in Appendix~\ref{appendix:pub_example}.

\textbf{Effect of the Refinement Loop.} Regarding the conventional clustering-based metrics, applying the feedback refinement loop does not result in significant performance differences. Specifically, no statistically significant differences are observed in P1, ARI, and NMI metrics for the Bills and Wiki datasets using a t-test under a p-value $<$ 0.05. In the Banking77 dataset, there is a a trend that the clustering-based metrics slightly decrease with more iterations, where the differences in P1 and ARI become statistically significant after 5 iterations. However, these are expected outcomes, as each iteration involves split and merge operations on only a small number of problematic topics. Additionally, the refinement process not only performs split and merge operations but also focuses on renaming topics—a factor that cannot be captured by clustering-based metrics. Consequently, most clusters remain unchanged, leading to minimal impact on clustering-based metrics.

On the contrary, the \ac{LLM}-as-a-Judge metrics show significant improvement, with the best performance achieved mostly with 3 refinement iterations. Both $\mathcal{TA}$ and $\mathcal{TC}$ exhibit statistically significant gains, as determined by a t-test with p-value $<$ 0.05, across all three datasets when using 3 iterations compared to results without the feedback refinement loop. Additionally, an interesting result is observed for the \ac{LLM}-ITL approach, which achieved relatively high P1 and NMI scores, as reported in its original paper. However, the $\mathcal{TA}$ and $\mathcal{TC}$ scores for \ac{LLM}-ITL are significantly lower than those of TIDE, TopigGPT, and \textsc{SeLATM}.\footnote{Also, the ARI score of \ac{LLM}-ITL is considerably low.} This suggests that, while a word-level \ac{LLM} analysis method like \ac{LLM}-ITL offers the advantage of reduced computational overhead, naming topics solely based on top-related words has the potential drawback of producing less accurate and less comprehensive topic names.

\begin{table}[t!]
    \caption{Experimental results on the business datasets.  We report an average of 5 repetitions. The best performance for each evaluation metric is highlighted in bold; underlined values indicate statistically significant performance gap over the best-performing baselines (or over \textsc{SeLATM} when a baseline performs best), determined by a t-test with a p-value $<$ 0.05.}
    \label{table:biz_data_experiments}
    \vspace{-1ex}
    \begin{center}
        \renewcommand{\arraystretch}{1.0}
        \footnotesize{
            \centering{\setlength\tabcolsep{2pt}}
        }
        \begin{tabular}{ccccccc}
        \toprule
        \multirow{2}{*}{\textbf{Models}} & \multicolumn{2}{c}{\textbf{CCC}} & \multicolumn{2}{c}{\textbf{BCR}} & \multicolumn{2}{c}{\textbf{CI}} \\
        & $\mathcal{TA}$ & $\mathcal{TC}$ & $\mathcal{TA}$ & $\mathcal{TC}$ & $\mathcal{TA}$ & $\mathcal{TC}$ \\ \midrule

        C-Top2vec & .723 & .886 & .050  & .016 & .607 &  .829 \\   
        TIDE & .750 & .773  & .502  & .555 & .836 & \underline{\textbf{.842}} \\   

        LLM-ITL  & .453 & .409 & .344 & .369 & .228 & .189 \\      
        \midrule

        \textsc{SeLATM} Iter:0 & .825 & .959 & .832 & .870 & .845 & .780 \\   

        \textsc{SeLATM} Iter:1 & .840 & .957 & .845  & .880 & \underline{\textbf{.847}} & .782 \\   

        \textsc{SeLATM} Iter:3 & \underline{\textbf{.849}} & \underline{\textbf{.962}} & .851  & .884 & .839 & .773 \\   

        \textsc{SeLATM} Iter:5 & .785 & .896 & \underline{\textbf{.870}} & \underline{\textbf{.899}} & .784 & .722 \\   
        \bottomrule
        \end{tabular}
    \end{center}
    \vspace{-3ex}
\end{table}

\begin{figure*}[ht]
    \centering
    \begin{subfigure}[b]{0.48\textwidth}
        \centering
        \includegraphics[width=\linewidth]{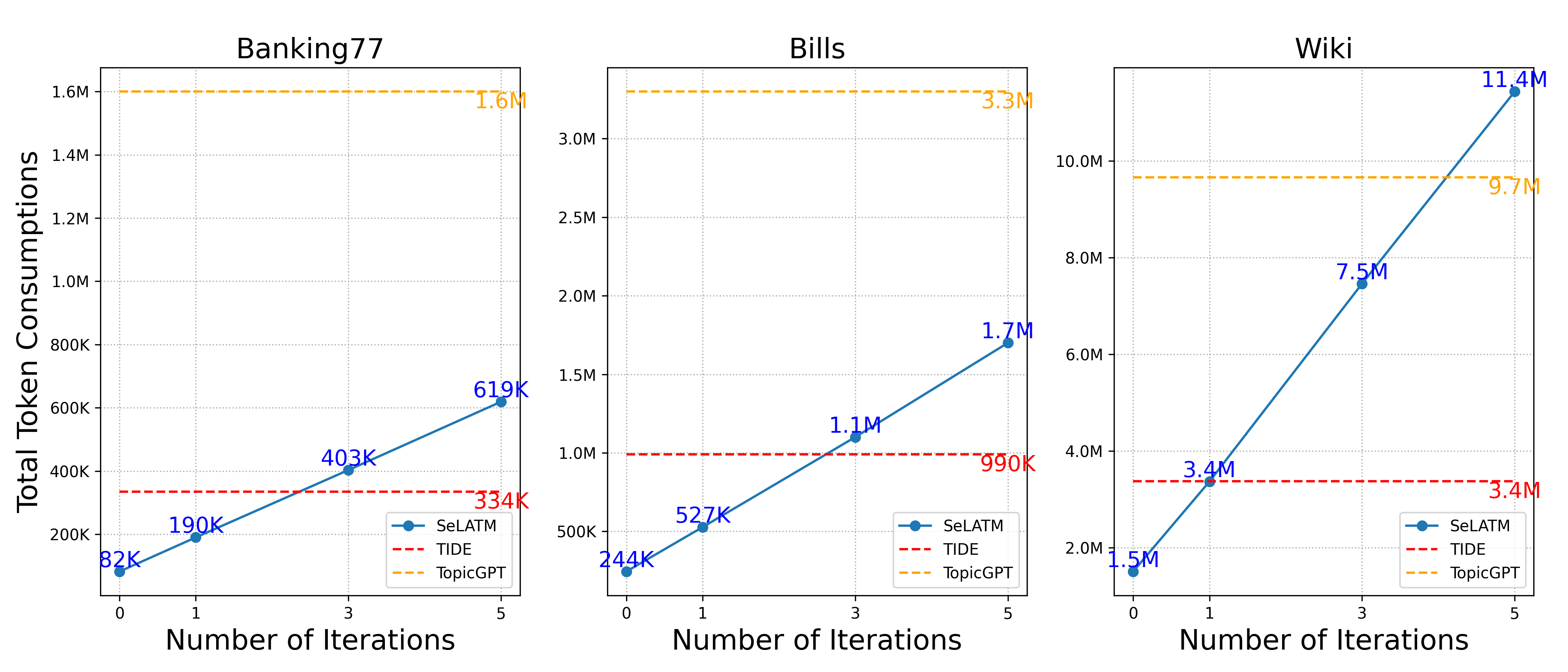}
        \caption{Public Datasets}
        \label{fig:pub_cost}
    \end{subfigure}
    \hspace{0.02\textwidth}
    \begin{subfigure}[b]{0.46\textwidth}
        \centering
        \includegraphics[width=\linewidth]{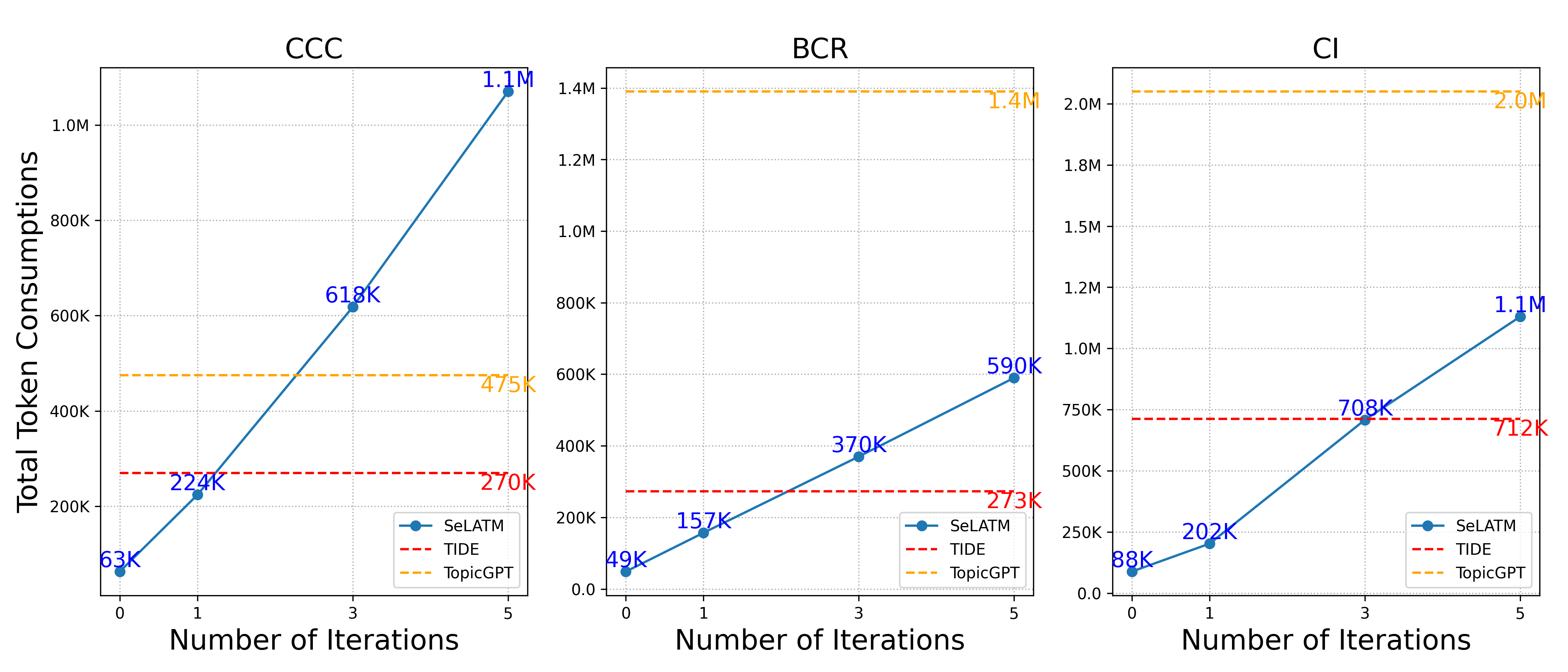}
        \caption{Business Datasets}
        \label{fig:biz_cost}
    \end{subfigure}
    \caption{An average LLM token consumption comparison between \textsc{SeLATM}, TIDE, and TopicGPT.}
    \label{fig:llm_cost_compare}
    \vspace{-3ex}
\end{figure*}

\subsection{Business Dataset Results}
For this experiment, we selected C-Top2vec, TIDE, and LLM-ITL as baselines. C-Top2vec and TIDE were chosen because they outperformed others within their respective topic modeling families by a large margin, while LLM-ITL was included as it is the sole representative of the \ac{LLM}-assisted neural topic model family. As the business datasets do not contain labels, only \ac{LLM}-as-a-Judge metrics are reported. The results are presented in Table~\ref{table:biz_data_experiments}. Several output examples of the business datasets are provided in Appendix~\ref{appendix:biz_example}.

Overall, the results exhibit a similar trend to those observed in the public dataset experiments. Even without the refinement loop, \textsc{SeLATM} statistically significantly outperforms baseline approaches in 5 out of 6  evaluation cases (based on a t-test with p-value $<$ 0.05), with the only exception being $\mathcal{TC}$ in the CI dataset. Additionally, \textsc{SeLATM} demonstrates consistently strong performance across all evaluation cases, in contrast to TIDE, which underperforms on the BCR dataset, and C-Top2vec, which performs poorly on $\mathcal{TA}$ for the CI dataset and fails entirely on the BCR dataset.\footnote{We observed that C-Top2vec generates only two topics on the BCR dataset, even with various hyperparameter settings.} The performance gap between baselines and \textsc{SeLATM} becomes even more pronounced with the feedback refinement. Specifically, the best results are achieved with 3 iterations for the CCC dataset and 5 iterations for the BCR dataset, with improvements that are statistically significant compared to \textsc{SeLATM} without the refinement loop, as determined by a t-test with p-value $<$ 0.05 (except for $\mathcal{TC}$ of the CCC dataset).

Although our proposed approach demonstrates notable improvements, we observed some limitations. First, the feedback refinement loop does not yield statistically significant improvement in the CI dataset. Additionally, we observed performance degradations after 5 iterations in both the CCC and CI datasets, which is quite similar to the results seen with the Banking77 dataset. Through qualitative analysis, we found that the split operation generally stops after 3 iterations, while the merge operation continues, inevitably resulting in larger but more ambiguous and generic topics, thereby leading to a decrease in the evaluation metrics. To address this issue, efforts should be made to adjust the judgment bias of $\mathcal{A}_{plan}$ so that excessive merging is recognized as a sign of negative change; we leave this as an area for future work. Despite these limitations, results from both public and business datasets indicate that \textsc{SeLATM} is likely to produce higher-quality topics than the baseline approaches, while also offering the benefit of greater interpretability through the quantification of contributions from multi-topics.

\begin{figure*}[ht]
    \centering
    \begin{subfigure}[b]{0.48\textwidth}
        \centering
        \includegraphics[width=\linewidth]{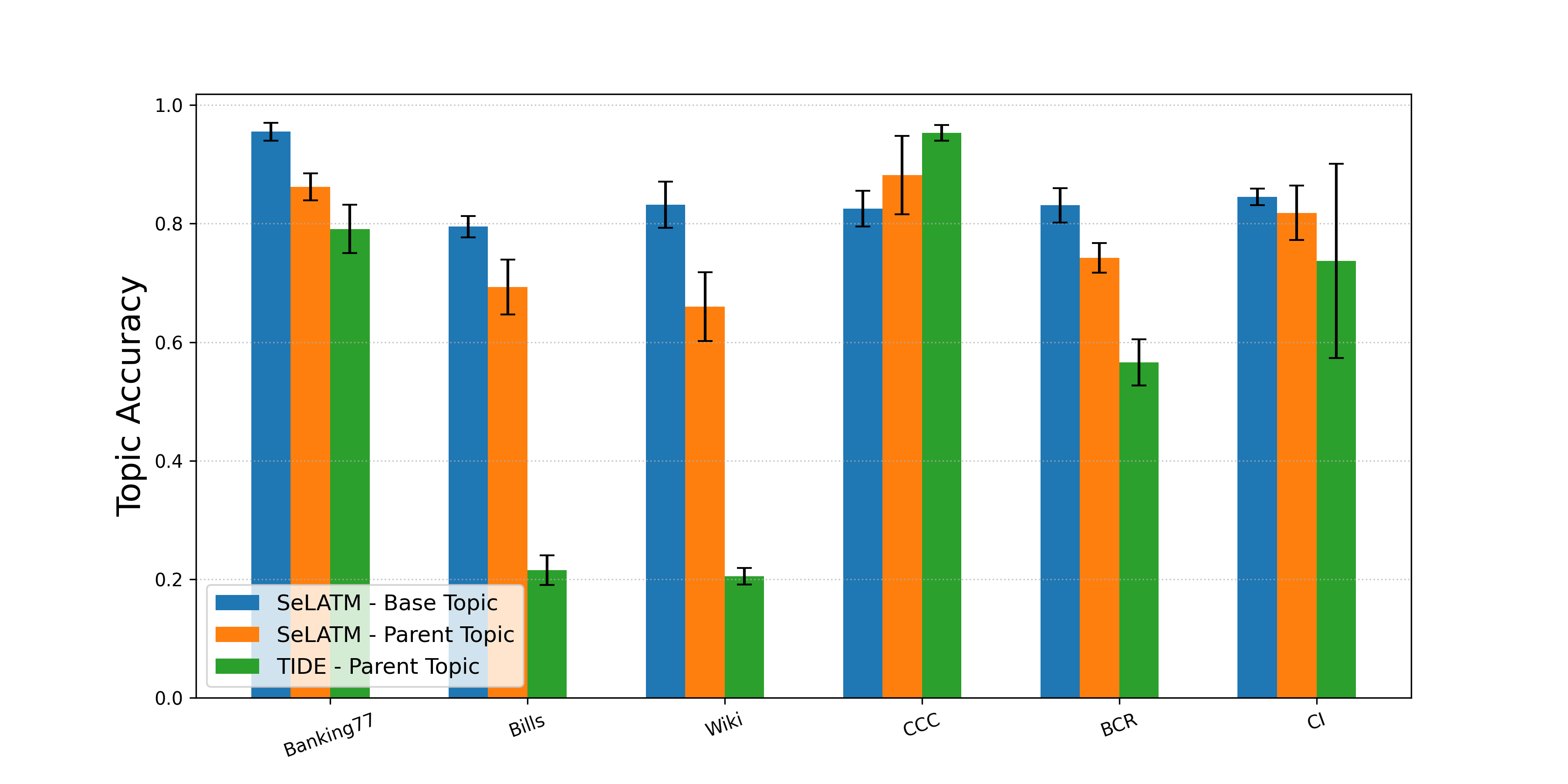}
        \label{fig:parent_topic_accuracy}
    \end{subfigure}
    \hspace{0.02\textwidth}
    \begin{subfigure}[b]{0.46\textwidth}
        \centering
        \includegraphics[width=\linewidth]{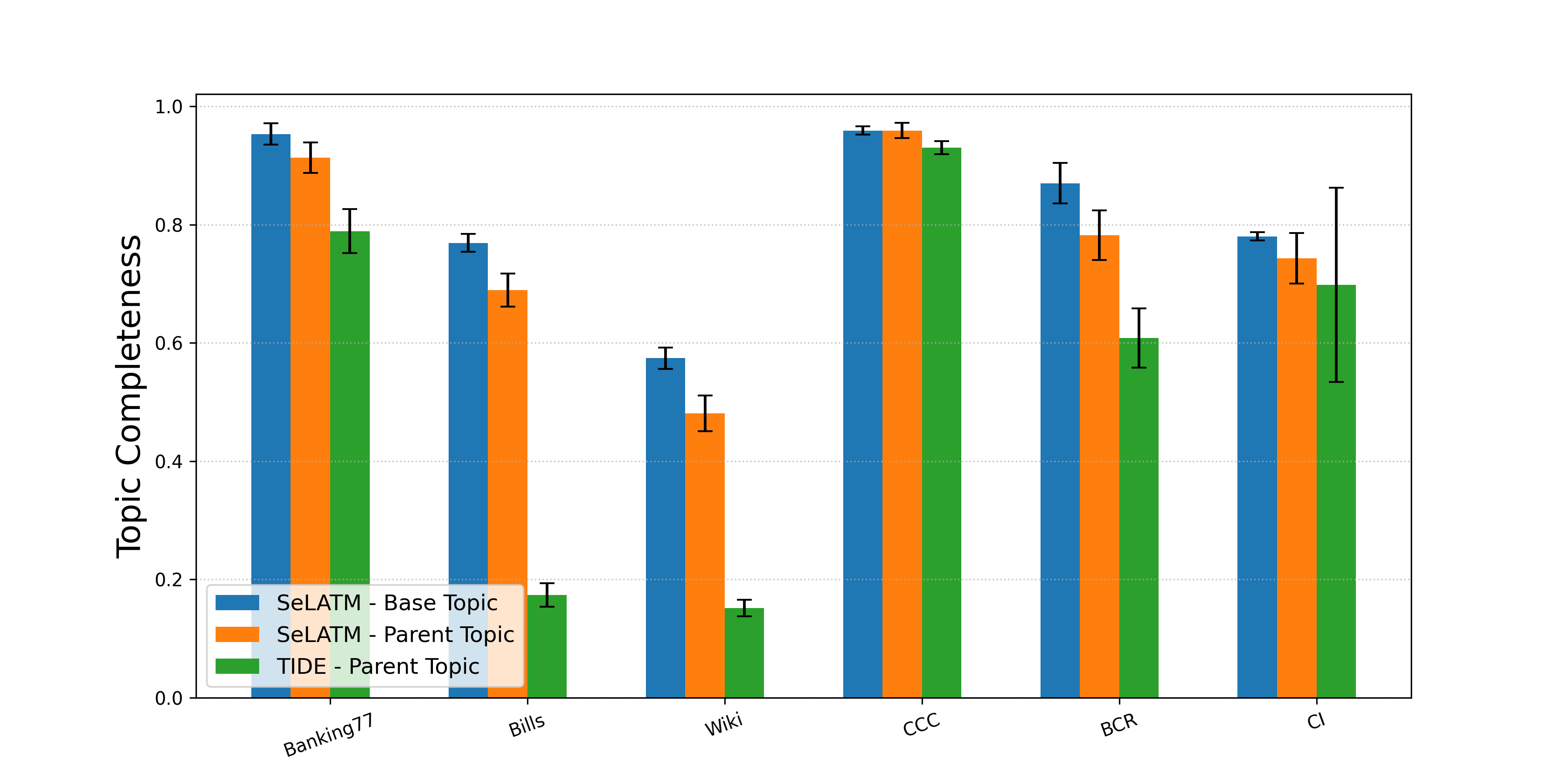}
        \label{fig:parent_topic_completeness}
    \end{subfigure}
    \vspace{-3ex}
    \caption{The $\mathcal{TA}$ and $\mathcal{TC}$ results on the parent topics.}
    \label{fig:parent_topic_results}
    \vspace{-3ex}
\end{figure*}

\subsection{LLM-Token Consumption Comparison}
As noted above, a key practical advantage of \textsc{SeLATM} over \ac{LLM}-driven topic modeling approaches that use topic generation followed by assignment is its \ac{LLM} resource efficiency. To verify this, we measured the total \ac{LLM} token consumption of the entire \textsc{SeLATM} process and compared them with that of TIDE and TopicGPT. LLooM is excluded from this comparison because its performance, as shown in Table~\ref{table:pub_data_experiments}, is significantly lower than that of the other methods in the same family. Figure~\ref{fig:llm_cost_compare} illustrates an average token consumption comparison. Without the feedback refinement loop, \textsc{SeLATM} uses significantly fewer \ac{LLM} resources—averaging only 24.4\% and 8.19\% of TIDE's and TopicGPT's token consumption, respectively. With the feedback refinement loop, \textsc{SeLATM} still uses only 55\% of TIDE's token consumption on average with one iteration, but begins to exceed TIDE's consumption after three iterations. Compared to TopicGPT, \textsc{SeLATM} uses fewer tokens even after 3 - 5 iterations. Given that \textsc{SeLATM} already outperforms TIDE and TopicGPT with zero or one iteration of the feedback refinement loop, as shown in Tables~\ref{table:pub_data_experiments} and \ref{table:biz_data_experiments}, the figures in Figure~\ref{fig:llm_cost_compare} highlight the superior resource efficiency of \textsc{SeLATM}.

\subsection{Parent Topic Modeling Performance}
To validate the quality of the parent topics, we measured \ac{LLM}-as-a-Judge evaluation metrics, as most of the datasets do not contain higher-level topic labels. We compared the performance of our proposed method with TIDE, which is the best-performing baseline capable of providing parent topics. The results are illustrated in Figure~\ref{fig:parent_topic_results}.

The results indicate that the parent topic achieves 92\% of the base topic's performance on average, which is notably high given that parent topics encompass a much broader range of content. Moreover, \textsc{SeLATM} produces significantly better parent topic performance compared to TIDE. This improvement is especially pronounced in the Bills and Wiki datasets, where \textsc{SeLATM} achieves $\mathcal{TA}$ and $\mathcal{TC}$ scores that are 2.2 and 2.6 times higher, respectively, than those of TIDE. For the other datasets, \textsc{SeLATM} improves performance by 11\% in $\mathcal{TA}$ and 14\% in $\mathcal{TC}$ compared to TIDE, with the only exception being the CCC dataset. TIDE achieves a higher $\mathcal{TA}$ in the CCC dataset; however, the difference is not statistically significant. The results highlight the advantage of multi-view clustering over fully relying on \ac{LLM} for identifying topic hierarchies.

Additionally, we conducted several ablation studies, which are presented in Appendix~\ref{appendix.ablation_study} due to the page limit constraint. We highly encourage readers to examine the ablation studies to investigate the benefits of using agglomerative clustering and the split operation.

\section{Related Works}
Topic modeling is an \ac{NLP} technique to automatically discover hidden semantic patterns underlying in a text corpus. Traditional statistical methods, such as \ac{LDA}~\cite{LDA} and \ac{NMF}~\cite{fevotte2011algorithms}, assume that topics are latent variables between words and documents, and attempt to uncover document-topic and topic-word distributions. A significant downside of these methods is that they rely on word-count based bag-of-words representations, which overlook the semantic meanings and relationships between words. In an attempt to overcome this drawback, embedded topic modeling approaches that utilize dense vector representations have been proposed~\cite{sia2020tired, top2vec, bertopic, contextual_top2vec}. The recent success of \acp{LLM} has enabled researchers to apply the models for topic modeling by directly asking \acp{LLM} to identify topics underlying a text corpus and assign them to each individual document~\cite{mu-etal-2024-large, doi-etal-2024-topic, topicgpt, concept_induction, LiSA}. A distinctive feature of \ac{LLM}-based methods is their capability to generate interpretable and human-readable topics, unlike the aforementioned approaches where topics are represented as a set of related words. However, these document-level \ac{LLM} analysis approaches have the drawback of significant computational overhead. To address this issue, LLM-ITL~\cite{LLM_ITL} utilizes conventional neural topic modeling for document analysis and employs \acp{LLM} primarily for refinement, thereby reducing computational overhead.

\section{Conclusion}
The advent and rapid progress of \acp{LLM} are propelling the field of topic modeling forward. Recent approaches actively employ \acp{LLM} to identify underlying topics within a corpus and assign them to individual documents. However, despite their promising results, recent \ac{LLM}-based topic modeling approaches have several drawbacks: (1) inability to quantify the contribution of multiple topics, (2) high computational demands due to document-level analysis, and (3) inaccurate topic naming resulting from insufficient refinement processes. In this paper, we propose \textsc{SeLATM} to address these limits, which is characterized by segment-level topic modeling and agentic feedback refinement loop. By assigning topics at the text segment level, \textsc{SeLATM} enables documents to be represented as distributions over multiple topics, thereby quantifying the contribution of each topic and elucidating the association between specific segments and their corresponding topics. Also, \textsc{SeLATM} eliminates the need for a separate topic assignment process, resulting in a substantial reduction in \ac{LLM} computational expense. Furthermore, both the segment-level topic modeling and feedback refinement loop  facilitate the generation of more precise topics by reducing task complexity and enabling collaborative refinement among diverse \ac{LLM} agents. Extensive experiments on public and business datasets demonstrate that \textsc{SeLATM} successfully addresses these challenges by enhancing interpretability through topic contribution quantification, reducing \ac{LLM} computation overhead, and producing better-quality topics.

\section*{Impact Statement}
This paper presents work whose goal is to advance the field of Machine
Learning. There are many potential societal consequences of our work, none
which we feel must be specifically highlighted here.

\begin{acronym}
  \acro{NLP}{Natural Language Processing}
  \acro{LLM}{large language model}
  \acro{PLM}{pre-trained language model}
  \acro{LDA}{Latent Dirichlet Allocation}
  \acro{NMF}{Non-negative Matrix Factorization}
\end{acronym}

\bibliography{references}
\bibliographystyle{icml2026}

\newpage
\appendix
\section{Appendix}

\subsection{Clustering-based Evaluation Metrics}~\label{appendix:cluster_eval_metrics}

Regarding the public datasets, we used clustering-based evaluation metrics in accordance with previous studies. Let us assume that there are $N$ instances and two clusters $U=\{u_1, u_2, ..., u_K\}$ and $V=\{v_1,v_2,...,v_C\}$, where $u_i$ and $v_i$ represent the $i$-th cluster assigned to $U$ and $V$, respectively.

\textbf{Harmonic Mean of Purity (P1).} Purity is an intuitive metric that is analogous to precision~\cite{zhao2001criterion}. Specifically, it evaluates how many ground-truth labels are contained within all documents assigned to a single predicted cluster. 
A small number of ground-truth labels in a single predicted cluster intuitively suggests a high degree of correspondence between the identified and ground-truth concepts~\cite{bills}. Provided $V$ to be considered as the ground-truth, purity is computed as follows:
\begin{align*}
    P(U,V)=\frac{1}{N} \sum_{k} \max_c |C_{U_k} \cap T_{V_c}|,
\end{align*}
where $C_{U_k}$ denotes the set of instances assigned to cluster $U_k$, and $T_{V_c}$ refers to the set of instances in ground-truth label $V_c$. However, purity is an unsymmetrical metric; therefore, the Harmonic Mean of Purity (P1) is more commonly used in practice, which is calculated as follows:
\begin{align*}
    P1(U,V)=\frac{2 \times P(U,V) \times P(V,U)}{P(U,V) + P(V,U)}.
\end{align*}

\textbf{Adjusted Random Index (ARI).}  Random Index (RI,~\citealt{RI}) is a metric that evaluates whether two clustering results are agree or disagree, which is calculated by:
\begin{align*}
    RI(U,V)= (N_{00} + N_{11})/{\binom{N}{2}},
\end{align*}
where $N_{00}$ and $N_{11}$ denote the number of pairs that are assigned to different clusters and same clusters, respectively, in both $U$ and $V$. However, the RI metric has certain limitations; for instance, although its theoretical range is from 0 to 1, its practical values often lie within the narrow range of 0.5 to 1~\cite{10.1145/1553374.1553511}. To address this, the ARI is more commonly used in practice, as it adjusts the RI to account for for the change groupings of elements~\cite{steinley2004properties}. The ARI is calculated as follows:
\begin{align*}
    ARI(U,V)=\frac{RI - RI_{\mathrm{E}}}{\max(RI) - RI_{\mathrm{E}}},
\end{align*}
where $RI_{\mathrm{E}}$ represents the expected value of the RI for random clusters, and $\max(R1)$ denotes the maximum possible value of the RI. The ARI ranges from -1 to 1, with 1 indicating perfect agreement between the two clusters, 0 corresponding to random cluster assignments, and negative values reflecting less agreement than would be expected by chance~\cite{ARI, 10.1145/1553374.1553511}.

\textbf{Normalized Mutual Information (NMI).} NMI is a metric used to evaluate the similarity between two clustering results, under the concept of mutual information (MI, \citealt{shannon1948mathematical}), which quantifies the amount of information shared between two random variables. As the name suggests, the NMI normalizes the MI to a value between 0 and 1, thereby reducing MI's sensitivity to variations in the number of clusters~\cite{strehl2002cluster}. The NMI is calculated as follows:
\begin{align*}
    NMI(U,V)=\frac{2 \times \mathcal{I}(U,V)}{\mathcal{H}_U + \mathcal{H}_V},
\end{align*}
where $\mathcal{I}(U,V)$ denotes the MI between the clusters $U$ and $V$. $\mathcal{H}_U$ and $\mathcal{H}_V$ refer to the entropies of the clusters $U$ and $V$, respectively.

\begin{table*}[t!]
    \caption{Results of $n_S$ search. We report an average of 5 repetitions. For each dataset, the best performance achieved by \textsc{SeLATM} with varying $n_S$ values is highlighted in bold. Regarding the Banking77 and BCR datasets, $n_S=80$ was not evaluated because the largest cluster in each dataset contained fewer than 80 instances. The best baseline performance is underlined if it achieves a higher score than the lowest performance of \textsc{SeLATM} across the different $n_S$ values.}
    \label{table:hps_search}
    \begin{center}
        \renewcommand{\arraystretch}{1.0}
        \footnotesize{
            \centering{\setlength\tabcolsep{2pt}}
        }
        \begin{tabular}{c cc cc cc cc cc cc}
        \toprule
        \multirow{2}{*}{\textbf{$n_S$}} & \multicolumn{2}{c}{\textbf{Banking77}} & \multicolumn{2}{c}{\textbf{Bills}} & \multicolumn{2}{c}{\textbf{Wiki}} & \multicolumn{2}{c}{\textbf{CCC}} & \multicolumn{2}{c}{\textbf{BCR}} & \multicolumn{2}{c}{\textbf{CI}} \\
        & $\mathcal{TA}$ & $\mathcal{TC}$ & $\mathcal{TA}$ & $\mathcal{TC}$ & $\mathcal{TA}$ & $\mathcal{TC}$ & $\mathcal{TA}$ & $\mathcal{TC}$ & $\mathcal{TA}$ & $\mathcal{TC}$ & $\mathcal{TA}$ & $\mathcal{TC}$ \\ \midrule

        30 & .932 & .924 & .775 & .744 & .814 & .586 & .791 & .954 & \textbf{.831} & \textbf{.870} & .753 & .703 \\
        40 & .935 & .927 & \textbf{.795} & \textbf{.769} & \textbf{.832} & .574 & .810 & .942 & .809 & .844 & .807 & .735 \\
        50 & .947 & .935 & .774 & .764 & .801 & .573 & \textbf{.825} & \textbf{.959} &.825 & .857 &  .796 & .723 \\
        60 & \textbf{.955} & .950 & .787 & .755 & .803 & .587 &  .812 & .958 &.806 & .839 & .832 & .762 \\
        70 & \textbf{.955} & \textbf{.953} & .777 & .734 & .805 & .584 & .821 & .958 &  .805 & .828 &\textbf{.845} & \textbf{.780} \\
        80 & - & - & .781 & .737 & .824 & \textbf{.603} & .826 & .958 & - & - & .806 & .747 \\
          
        \midrule
        Best Baseline Result & .782 & .867 & .759 & .719 & .641 & \underline{.621} & .750 & .886 & .502 & .555 & \underline{.836} & \underline{.842} \\   

        \bottomrule
        \end{tabular}
    \end{center}
    \vspace{-1ex}
\end{table*}

\subsection{Baseline Models and Implementation Details}~\label{appendix.baselines}

\textbf{BERTopic}~\cite{bertopic}: BERTopic belongs to an embedded topic models family that employs document embeddings derived from a \ac{PLM}. The embeddings are then clustered, and for each cluster, topic representations are generated using a class-based TF-IDF method. For implementation, we used the public BERTopic library.\footnote{\href{https://github.com/MaartenGr/BERTopic}{https://github.com/MaartenGr/BERTopic}} In accordance with the paper, we employed \texttt{all-mpnet-base-v2} as a backbone embedding model and used the $fit$ API with its default configurations.
    
\textbf{C-Top2Vec}~\cite{contextual_top2vec}: This method is more improved embedded topic model compared to BERTopic. It segments a document using a sliding window and represents it within multidimensional embedding vector representations. The vector representations are clustered, with topic vectors defined as the centroids of these clusters. Document-topic distributions are then determined according to the topics to which the segments are assigned. Finally, each topic is labeled using words and phrases that are proximal to the corresponding topic vectors. For implementation, we used the original code shared by the authors.\footnote{\href{https://github.com/ddangelov/Top2Vec}{https://github.com/ddangelov/Top2Vec}} Similar to the paper, we employed \texttt{all-mpnet-base-v2} as an embedding model and applied the same hyperparameter settings as the authors, which are demonstrated in Table 8 of their paper~\cite{contextual_top2vec}.

\textbf{TopicGPT}~\cite{topicgpt}: TopicGPT is an \ac{LLM}-based approach comprising two stages: topic generation, which identifies the underlying topics within a given corpus, and topic assignment, which allocates these generated topics to individual documents For implementation, we used the public code shared by the authors.\footnote{\href{https://github.com/chtmp223/topicGPT}{https://github.com/chtmp223/topicGPT}} For experiments, we used the full pipeline including \texttt{refinement}. To make fair comparison with other methods that generates more granular topics, we generated and evaluated on low-level topic produced by $generate\_topic\_lvl2$ API. 

\textbf{LLoom}~\cite{concept_induction}. This \ac{LLM}-based framework is designed to identify high-level concepts. The method involves three steps: \textsc{Distill}-\textsc{Cluster}-\textsc{Synthesize} steps. The \textsc{Distill} step condenses text data through summarization. The \textsc{Cluster} step performs clustering and \textsc{Synthesize} step identifies high-level concepts underlying in each cluster. The \textsc{Distill} and \textsc{Synthesize} steps are \acp{LLM}-driven processes. For implementation, we used the official code shared by the authors.\footnote{\href{https://github.com/michelle123lam/lloom}{https://github.com/michelle123lam/lloom}} For all of our experiments, We utilized the full pipeline exposed by $gen\_auto$ API, using the default parameter settings. 

\textbf{LLM-ITL}~\cite{LLM_ITL}. This approach utilizes \acp{LLM} for topic refinement. Specifically, topics—represented as word distributions—and document representations—expressed as topic distributions—are first learned through neural topic modeling (NTM). Subsequently, an \ac{LLM} generates the topic name based on the topic word distribution and refines the topic words by removing irrelevant terms and adding new, relevant ones. A distinctive feature of \ac{LLM}-ITL is its use of Optimal Transport (OT) distance to optimally align the topic word distributions produced by the neural topic model (NTM) with the refined topic word distributions generated by the \ac{LLM}. The approach has several advantages in that it can be readily applied to various NTM methods and significantly reduces \ac{LLM} usage overhead. For implementation, we used the original code shared by the authors.\footnote{\href{https://github.com/Xiaohao-Yang/LLM-ITL}{https://github.com/Xiaohao-Yang/LLM-ITL}} We utilized NVDM~\cite{miao2017discovering} as the base NTM method and adopted the same training hyperparameters as those used by the authors, except for the number of topics, which was set to match our experimental setup.

\textbf{TIDE}~\cite{moon2026industryalignedgranulartopicmodeling}. This approach is a modified version of TopicGPT. Unlike TopicGPT, which generates topics from entire documents, TIDE first performs clustering and then generates topics for each cluster, enabling to use fewer \ac{LLM} calls. We introduced TIDE as a baseline because it enables users to decide the number of topics, allowing a fairer comparison with LLM-ITL and \textsc{SeLATM}. Although the code is not publicly available, the paper provides a clear description of the algorithm and detailed information about the prompt designs. Furthermore, the algorithm is straightforward to replicate; therefore, we implemented it ourselves by referring to the paper.

\begin{table*}[t!]
    \caption{Ablation study on the clustering methods. We report an average of 5 repetitions (Avg.) and their standard deviations (Std.). The best average performance for each evaluation metric is highlighted in bold. The best performance exhibits statistically significant gap (*) determined by t-test with a p-value $<$ 0.05.}
    \label{table:ablation_clustering}
    \vspace{-1ex}
    \begin{center}
        \renewcommand{\arraystretch}{1.0}
        \footnotesize{
            \centering{\setlength\tabcolsep{1pt}}
        }
        \begin{tabular}{ccccccccccc}
        \toprule
        & \multirow{2}{*}{\textbf{Models}} & \multicolumn{3}{c}{\textbf{Banking77}} & \multicolumn{3}{c}{\textbf{Bills}} & \multicolumn{3}{c}{\textbf{Wiki}} \\ & 
        & P1 & ARI & NMI & P1 & ARI & NMI & P1 & ARI & NMI \\ \midrule

        \multirow{2}{*}{\textbf{Avg.}} &
        K-Means & .664 & .524 & .819 & .415 & .247 & \textbf{.680}* & .500 & .334 & .788\\
       &  Agglomerative & .\textbf{684}* & \textbf{.542}* & \textbf{.832}* & \textbf{.440}* & \textbf{.255} & .663 & \textbf{.522}* & \textbf{.357}* & \textbf{.805}* \\ \midrule

        \multirow{2}{*}{\textbf{Std.}} &
        K-Means & .014 & .018 & .007 & .011 & .026 & .005 & .012 & .023 & .004 \\
        &  Agglomerative & .002 & .002 & .000 & .005 & .002 & .001 & .004 & .005 & .001 \\

        \bottomrule
        \end{tabular}
    \end{center}
    \vspace{-3ex}
\end{table*}

\subsection{Hyperparameter Search for Optimal $n_S$}~\label{appendix.hpo_Ns}
In \textsc{SeLATM}, $n_S$ is a crucial hyperparameter that determines the number of text segment samples for each cluster used for topic generation and, consequently, may influence the quality of the generated topics, e.g., the name and description of topics. Accordingly, we conducted a thorough hyperparameter search across all datasets, varying the value from 30 to 80 in increments of 10. We measured $\mathcal{TA}$ and $\mathcal{TC}$ for the hyperparameter search, as the clustering-based metrics are already determined during segmentation and clustering phase and do not depend on the sampled text segments. The results are summarized in Table~\ref{table:hps_search}.

Based on the search result, we set $n_S$ to 70 for the Banking77 and CI datasets, 40 for the Bills and Wiki datasets, 50 for the CCC dataset, and 30 for the BCR dataset. However, the effect of $n_S$ was less pronounced than expected, with only marginal performance differences across most datasets, except for the CI dataset. Moreover, even the lowest performance achieved by \textsc{SeLATM} with specific $n_S$ exceeds that of the best baseline approach in 9 out of 12 evaluation cases. Therefore, we recommend setting $n_S$ to around 40–50 if there is insufficient capacity to conduct a thorough hyperparameter search.

\subsection{Ablation Study}~\label{appendix.ablation_study}

In this section, we conduct ablation studies to ascertain the benefit of utilizing agglomerative clustering and the \texttt{split} operation.

\noindent\textbf{Comparison with K-Means Clustering}. A leading cause of adopting agglomerative clustering is because the approach is more stable than K-Means clustering, which tends to have greater variability. As part of our ablation study, we evaluated and compared the performance of K-Means clustering against agglomerative clustering. Specifically, the experiments were conducted on public datasets, and clustering-based evaluation metrics were used to assess performance. No feedback refinement loop is applied. The results are summarized in Table~\ref{table:ablation_clustering}.

The findings demonstrate that agglomerative clustering exhibits greater stability and achieves superior evaluation metric scores compared to K-Means. With the exception of the Bills dataset, agglomerative clustering demonstrates statistically significantly superior performance compared to K-Means across all evaluation metrics. While K-Means achieves a marginally higher NMI score on the Bills dataset, agglomerative clustering surpasses it on the other two metrics. In terms of standard deviations, agglomerative clustering exhibits substantially lower variability compared to K-Means, which further support our choice of using agglomerative clustering for this study. M

\begin{table*}[t!]
    \caption{Ablation study on the \texttt{split} operation. We report an average of 5 repetitions. The performance of `\texttt{split}+\texttt{merge}' is underlined when it is statistically significantly higher than that of `w/o \texttt{split}' in the corresponding iteration under a t-test with a p-value $<$ 0.05.}
    \label{table:ablation_split}
    \vspace{-1ex}
    \begin{center}
        \renewcommand{\arraystretch}{1.0}
        \footnotesize{
            \centering{\setlength\tabcolsep{1pt}}
        }
        \begin{tabular}{cccccccc}
        \toprule
        & \multirow{2}{*}{\textbf{Iterations}} & \multicolumn{2}{c}{\textbf{Banking77}} & \multicolumn{2}{c}{\textbf{Bills}} & \multicolumn{2}{c}{\textbf{Wiki}} \\ & 
        & $\mathcal{TA}$ & $\mathcal{TC}$ & $\mathcal{TA}$ & $\mathcal{TC}$ & $\mathcal{TA}$ & $\mathcal{TC}$ \\ \midrule

        no Iteration & 0 & .955 & .953 & .795 & .769 & .832 & .574 \\ \midrule

        \multirow{3}{*}{\texttt{split} + \texttt{merge}} &
        1 & .962 & .956 & \underline{.803} & \underline{.782} & \underline{.807} & \underline{.567} \\
        & 3 & .966 & \underline{.960} & \underline{.823} & \underline{.792} & \underline{.842} & \underline{.587} \\
        & 5 & .962 & .961 & \underline{.815} & \underline{.789} & \underline{.841} & \underline{.588} \\ \midrule

        \multirow{3}{*}{w/o \texttt{split}} &
        1 & .957 & .955 & .754 & .770 & .687 & .525\\
        & 3 & .959 & .949 & .755 & .751 & .639 & .499 \\
        & 5 & .962 & .958 & .727 & .728 & .618 & .481 \\

        \bottomrule
        \end{tabular}
    \end{center}
\end{table*}

\textbf{Effect of Split Operation}. One distinctive feature of our proposed method's refinement process is the inclusion of a \texttt{split} operation, whereas most existing \ac{LLM}-driven topic modeling approaches rely solely on topic merging. In order to assess the impact of the \texttt{split} operation, we conducted an ablation study in which $\mathcal{E}_{coh}$ and $\mathcal{A}_{split}$ were removed from the refinement loop, so that only $\mathcal{E}_{div}$ and $\mathcal{A}_{merge}$ were utilized. The experiments were conducted on the public datasets, and \ac{LLM}-as-a-Judge metrics were measured, since the feedback refinement loop has only a minor impact on clustering-based metrics but a significant effect on \ac{LLM}-as-a-Judge metrics. The results are summarized in Table~\ref{table:ablation_split}.

In the Banking77 dataset, the inclusion of the \texttt{split} operation does not result in any significant differences. However, considerable performance degradations are observed in the Bills and Wiki datasets, where the absence of the \texttt{split} operation consistently resulted in statistically significantly lower performance across all iterations compared to the full feedback refinement loop. Moreover, in both datasets, performance continues to decline with additional iterations when the \texttt{split} operation is omitted, which is an expected outcome since excessive topic merging leads to broader and more ambiguous topics. Additionally, we observed that the \texttt{split} operation ceases to be applied after 2 to 3 iterations, which is closely related to why optimal performance is typically achieved within 1 to 3 iterations. These findings substantiate our claim regarding the necessity of including the \texttt{split} operation in the refinement process.

\section{Prompt Designs}
\subsection{Topic Generation Prompt}~\label{appendix.topic_gen_prompt}
To generate topic names and descriptions of a given text segment cluster, we use the following prompt, where \texttt{n\_topic\_words} denotes a user-defined hyperparameter with a default value of 2. \texttt{topic\_definition} is an optional prompt that allows users to specify a customized definition of a topic, which is particularly useful when working with real-world industry data.

\begin{tcolorbox}[
  title={Prompt for Topic Generation Agent},
  colback=gray!5,
  colframe=black,
  fonttitle=\bfseries,
  breakable
]

\medskip
You will receive a list of text segments that belong to the same cluster. Each text segment is identified by a unique ID. Given the segments, your task is to read the contents, then generate a topic that can describe as many text segments from this group as possible.

Along with the topic, you need to generate a short description of the meaning of the topic given by yourself.

\medskip

\textbf{[Instructions]}

\{\% if \{n\_topic\_words\} $<$ 1\%\}

- The topic should be a short phrase of \{\{n\_topic\_words-1\}\} - \{\{n\_topic\_words\}\} words.

\{\% endif \%\}

\{\% if \{n\_topic\_words\} == 1\%\}

 - The topic should be a single word.

\{\% endif \%\}

- The description should be no more than two sentences.

 - The description needs to start with phrase `This topic is about...'

\medskip

\{\% if \{topic\_definition\} is  not none \%\}

\{\{topic\_definition\}\}

\{\% endif \%\}
\medskip

\{\% if (\{previous\_topic\} is not none) and (\{feedback\} is not none) \%\}

This cluster was previously named as \{\{previous\_topic\}\}. Here is some feedback on how you can improve the quality of the topic: \{\{feedback\}\}
\{\% endif \%\}

\medskip
Please keep in mind that the text segments with different ID does not originated from the same document.

\medskip
\textbf{Text Segments}: \{\{text\_segments\}\}
\end{tcolorbox}

The response is generated using structured \ac{LLM} generation, following the \texttt{pydantic} schema shown below.

\begin{lstlisting}
from pydantic import BaseModeltopic_eval_coh

class TopicGenerationOutput(BaseModel)
    topic: str
    description: str
\end{lstlisting}

\subsection{Topic Coherence Evaluation Prompt}~\label{appendix.topic_eval_coh}
We evaluate the coherence of each generated topic, which measures how closely the topic name and description match with the assigned text segments. The prompt presented below is used for this.

\begin{tcolorbox}[
  title={Prompt for Topic Coherence Evaluation Agent},
  colback=gray!5,
  colframe=black,
  fonttitle=\bfseries,
  breakable
]

\medskip
You’ll receive a list of text segments, a topic assigned to the given corpus, and its corresponding description.

Your task is to calculate the Topic Coherence score, which evaluates how closely the topic name and the description match with the text segments.

\medskip
Assign an overall Topic Coherence score from 1 to 10 (including .5 increments), where:

- 10 = High topic coherence: The topic name and description align very well with the text segments.

- 1 = Low topic coherence: The topic name and description covers too broadly compared to the text segments.

\medskip
If the topic coherence score is sufficient and does not require additional separation:

- specify the field `need\_separation' as False.

- leave the fields `feedback' as null.

\medskip
If the topic coherence score is low and requires separation, please:

- Specify the name of the problematic topic in the `topic' field.

- Specify the field `need\_separation' as True.

- Provide detailed feedback in the `feedback' field by addressing the following points:

    - Explain the aspects of the topic and description that do not align well with the text segments.
    
    - Offer concrete suggestions to improve the coherence between the topic and the text segments.

\medskip
\textbf{List of Text Segments}: \{\{text\_segments\}\}

\textbf{Assigned Topic}: \{\{topic\_name\}\}

\textbf{Topic Description}: \{\{topic\_description\}\}
\end{tcolorbox}

The response is generated using structured \ac{LLM} generation, following the \texttt{pydantic} schema shown below.

\begin{lstlisting}
from pydantic import BaseModel

class TopicCoherenceEvalIndividualOutput(BaseModel):
    topic: str
    score: float
    need_separation: bool
    feedback: str | None
\end{lstlisting}

\subsection{Topic Diversity Evaluation Prompt}~\label{appendix.topic_eval_div}
Topic diversity evaluation agent performs a holistic topic evaluation, which measures the degree of distinctness among topics—that is, the extent to which they are dissimilar from one another. The prompt presented below is used for this task.

\begin{tcolorbox}[
  title={Prompt for Topic Diversity Evaluation Agent},
  colback=gray!5,
  colframe=black,
  fonttitle=\bfseries,
  breakable
]

\medskip
You will receive a list of topics, which are identified by a unique ID, and their corresponding descriptions.

Your task is to calculate the Topic Diversity score, which assesses how distinct the topics are from one another and helps identify if any topics need to be merged or are too similar.

\medskip
Assign an overall Topic Diversity score from 1 to 10 (including .5 increments), where:

- 10 = High topic diversity: The topics are very distinct from one another.

- 1 = Low topic diversity: Some topics are very similar and may need to be merged.

\medskip
If the topic diversity score is sufficient and does not require merging:

- specify the field `need\_merge' as False.

- leave the fields `feedback' and `topics\_to\_merge' as null.

\medskip
If the topic diversity score is low and requires merging, please:

- Specify the field `need\_merge' as True.

- Provide detailed feedback in the `feedback` field by addressing the following points:

    - Explain which topics are too similar and why they should be merged.
    
    - Offer concrete suggestions on how to improve topic diversity by merging similar topics.

- For each group of similar topics, list the IDs of the TWO most similar topics to be merged in the `topics\_to\_merge' field.

\medskip
\textbf{Topics and Descriptions}: \{\{topic\_results\}\}
\end{tcolorbox}

The response is generated using structured \ac{LLM} generation, following the \texttt{pydantic} schema shown below.

\begin{lstlisting}
from pydantic import BaseModel

class TopicDiversityEvalOutput(BaseModel):
    score: float
    need_merge: bool
    feedback: str | None
    topics_to_merge: list[list[int]] | None
\end{lstlisting}

\subsection{Planning Agent Prompt}~\label{appendix.planner_prompt}
The planner agent first generates summary reports for both before and after the topic refinement iteration by taking the coherence and diversity evaluation results as input and using the following prompt:

\begin{tcolorbox}[
  title={Prompt for Summary Report Generation},
  colback=gray!5,
  colframe=black,
  fonttitle=\bfseries,
  breakable
]

\medskip
You will receive a list of topics along with feedback, highlighting areas that need improvement.

Your task is to create an evaluation report that summarizes this feedback.

\medskip
\textbf{[Instruction]}

- Analyze the feedback: Carefully review the comments from evaluators on both topic diversity and topic coherence. Identify key strengths and weaknesses highlighted in the feedback.

- Generate an evaluation report: Based on your analysis, create a concise report that highlights the strengths and weaknesses of the topic modeling results.

- Keep in mind not to include suggestions that are not described in the given feedback.

\medskip Please include an overall score in the `score` field, ranging from 1 to 10, based on the strengths, weaknesses, and individual scores mentioned in the feedback.

\medskip
\textbf{Coherence Evaluation Results}: \{\{topic\_coherence\_eval\_results\}\}

\textbf{Diversity Evaluation Results}: \{\{topic\_diversity\_eval\_results\}\}
\end{tcolorbox}

The response is generated using structured \ac{LLM} generation, following the \texttt{pydantic} schema shown below.

\begin{lstlisting}
from pydantic import BaseModel

class PlannerEvalReportOutput(BaseModel):
    coherence_score: float
    diversity_score: float
    report_summary: str
\end{lstlisting}

Once the summary reports are generated, the planning agent determines whether improvements have been achieved by comparing the two. The task is conducted by employing the following prompt:

\begin{tcolorbox}[
  title={Prompt for Improvement Check},
  colback=gray!5,
  colframe=black,
  fonttitle=\bfseries,
  breakable
]

\medskip
You will receive two topic modeling evaluation reports: one from before and one from after the current refinement iteration.

Your task is to determine whether there has been an improvement in the topic modeling results based on these reports.

\medskip
\textbf{[Instruction]}

For making your decision, consider the following aspects:

1) Overall Score Comparison: Compare the overall scores from both reports. An improvement is indicated if the post-refinement report has a higher overall score than the pre-refinement report.

2) Strengths and Weaknesses: Analyze the strengths and weaknesses highlighted in both reports. Look for areas where the post-refinement report shows fewer weaknesses or improved strengths compared to the pre-refinement report.

3) Number of Topics that Need Refinement: Count the number of topics that were identified as needing refinement in both reports. An improvement is indicated if the post-refinement report has fewer topics requiring refinement than the pre-refinement report.

4) Specific Feedback Analysis: Review the specific feedback provided for each topic in both reports. Check how severe the issues were in the pre-refinement report and whether they have been adequately addressed in the post-refinement report. Look for positive changes or resolved issues in the post-refinement report compared to the pre-refinement report.

\medskip 
If the improvement criteria are met, indicate `is\_improved` as `True`. Otherwise, indicate `is\_improved` as `False`. Also, provide a brief explanation of your decision in `reason` field.

\medskip
\textbf{Pre-refinement Report}: \{\{report\_before\}\}

\textbf{Post-refinement Report}: \{\{report\_after\}\}
\end{tcolorbox}

The response is generated using structured \ac{LLM} generation, following the \texttt{pydantic} schema shown below.

\begin{lstlisting}
from pydantic import BaseModel

class PlannerImprovementCheckOutput(BaseModel):
    is_improved: bool
    reason: str
\end{lstlisting}

\subsection{LLM-as-a-Judge Prompt}~\label{appendix.llm_as_judge_prompt}
To evaluate topic accuracy ($\mathcal{TA}$) and topic completeness ($\mathcal{TC}$) using \ac{LLM}-as-a-Judge, we used the following prompts:

\begin{tcolorbox}[
  title={Prompt for Topic Accuracy Evaluation},
  colback=gray!5,
  colframe=black,
  fonttitle=\bfseries,
  breakable
]
\medskip
We will show you a model-generated topics, which are assigned to the given text.

\medskip
Your task is to evaluate the accuracy of the generated topic topic over the given text; how accurate the topic is and well represented in the given text.

\medskip
You can ignore the level of granularity / detail. Please just assess if the topic is accurate given the text in any level of detail.

\medskip
You should evaluate the topic accuracy in four level: Incorrect, Partially Correct, Mostly Correct, and Completely Correct. The definitions of each level are as follows:

- Incorrect: The details in the topic is not represented in the text in any way. The topic label is completely wrong or directly contradicting the text.

- Partially Correct: The details of the topic is partially represented in the text. There is some relevance of the topic label to the contents in the text.

- Mostly Correct : The details of the topic is mostly represented in the text. 

- Completely Correct : All the details of the topic are represented in the text. 

\medskip
\textbf{Text}: \{\{text\}\}

\textbf{Assigned Topic}: \{\{topic\}\}
\end{tcolorbox}

\begin{tcolorbox}[
  title={Prompt for Topic Completeness Evaluation},
  colback=gray!5,
  colframe=black,
  fonttitle=\bfseries,
  breakable
]

\medskip
We will show you a model-generated topic(s), which are assigned to the given text.

\medskip
Your task is to evaluate the completeness of the topic, which assesses whether there exist any other topics that are present in the given text but missing.

\medskip
You should evaluate the topic completeness in four level: Not Covered, Minorly Covered, Mostly Covered, and Complete. The definitions of each level are as follows:

- Not Covered: The assigned topic(s) do not reflect any of the main themes or subjects present in the text. Significant topics are missing, and the model’s output fails to capture the core content.

- Minorly Covered: The assigned topic(s) capture only a small portion of the relevant themes in the text. Most important topics are missing, and the coverage is insufficient to represent the text’s content.

- Mostly Covered: The assigned topic(s) capture most of the main themes or subjects in the text, but one or more significant topics are still missing. The coverage is substantial but not fully complete.

- Complete: The assigned topic(s) fully reflect all major themes and subjects present in the text. No significant topics are missing, and the model’s output provides a comprehensive representation of the text.

\medskip
\textbf{Text}: \{\{text\}\}

\textbf{Assigned Topic}: \{\{topic\}\}
\end{tcolorbox}

The responses of both evaluation dimensions are generated using structured \ac{LLM} generation, following the \texttt{pydantic} schema shown below.

\begin{lstlisting}
from pydantic import BaseModel

class AutoEvalOutput(BaseModel):
    answer: str
    reason: str
\end{lstlisting}

\clearpage
\onecolumn
\section{Examples}
\subsection{Public Dataset Examples}~\label{appendix:pub_example}

\begin{table}[h]
    \caption{Examples of public datasets and their corresponding topics generated by \textsc{SeLATM}.}
    \begin{center}
    \renewcommand{\arraystretch}{1.5}
    \footnotesize{
        \centering{\setlength\tabcolsep{2.0pt}}
    }

    \begin{tabular}{>{\centering\arraybackslash}m{2.0cm}>{\arraybackslash}m{10.5cm}>{\arraybackslash}m{3.0cm}}
        \toprule
        \textbf{Dataset} & \makecell[c]{\textbf{Input Text}} & \makecell[c]{\textbf{Generated Topic}} \\ \midrule
        Banking77 & The exchange rate applied is wrong for my international purchase.	& \{`Incorrect Exchange Rates': 1.0\} \\

        Bills & Restore American Dream Act of 2007 - Amends the Internal Revenue Code to establish tax-exempt homeownership plans. Allows a tax deduction from gross income for cash contributions to such plans. Defines "homeownership plan" as a trust established for the exclusive purpose of paying the costs (e.g., downpayment, interest, mortgage insurance, closing costs, etc.) of acquiring a principal residence by an individual who has never owned a principal residence. Excludes from gross income distributions from a homeownership plan used to pay the costs of acquiring a principal residence. Sets forth rules governing: (1) transfers of an interest in a plan due to death or divorce; (2) penalties for making distributions from a plan for purposes other than to acquire a principal residence; and (3) employer homeownership plans established for the benefit of their employees.	 & \{`Housing Legislation': 0.2, `Tax Legislation': 0.8\} \\ 

        Wiki &

        Borodino-class battlecruiser = The Borodino-class battlecruisers were a group of four battlecruisers ordered by the Imperial Russian Navy before World War I. Also referred to as the Izmail class , they were laid down in December 1912 at Saint Petersburg for service with the Baltic Fleet. [...]  None of the proposals was accepted, and all three of the less complete ships were sold to a German company for scrap on 21 August 1923 to raise much-needed cash for the government . The Operational Administration of the Soviet Navy worked out the requirements in May 1925 for a conversion that would have made Izmail into an aircraft carrier with a top speed of 27 knots (50 km / h;31 mph) and a capacity of fifty aircraft. She would have been armed with eight 183-millimetre (7.2 in) guns and her armour reduced to a maximum of 76 millimetres (3.0in). This proposal was approved by Alexey Rykov , Chairman of the Council of the People 's Commissars on 6 July 1925. The plan was cancelled on 16 March 1926 after the Red Army managed to gain control of a commission appointed to review the needs of the Navy, and the Army was strongly opposed to spending additional money on naval projects. She was scrapped beginning in 1931 in Leningrad . = = Ships = = The ships were named after battles fought by the Russian Empire: &
        \{`Naval Warfare': 0.517, `Naval Armament': 0.379, `Naval Propulsion': 0.103\} \\
        
    \bottomrule
    \end{tabular}
    \end{center}
    \label{tab:example_pub}
\end{table}

\begin{table}[h]
    \caption{The descriptions of topics presented in Table~\ref{tab:example_pub}.}
    \begin{center}
    \renewcommand{\arraystretch}{2.0}
    \footnotesize{
        \centering{\setlength\tabcolsep{1.5pt}}
    }

    \begin{tabular}{>{\arraybackslash}m{3.5cm}>{\arraybackslash}m{12.0cm}}
        \toprule
        \makecell[c]{\textbf{Topic}} & \makecell[c]{\textbf{Generated Topic Description}} \\ \midrule
        Incorrect Exchange Rates & This topic is about issues related to incorrect exchange rates applied during purchases or cash withdrawals in foreign currencies. Customers are expressing concerns about discrepancies between expected and actual exchange rates applied to their international transactions. \\

        Housing Legislation & This topic is about various legislative acts and amendments related to housing policies, including public housing, tax exemptions for homeownership, mortgage regulations, and housing assistance programs. It encompasses issues such as rent adjustments, affordable housing eligibility, support for vulnerable populations, and regulatory simplifications to improve housing access and management. \\

        Tax Legislation & This topic is about various legislative amendments to the Internal Revenue Code aimed at modifying tax rates, exclusions, and rules related to corporate, estate, and investment taxation. These amendments address issues such as estate and gift tax exclusions, corporate expatriations, timber gains, and regulated investment companies, reflecting ongoing efforts to refine tax policies and their economic implications. \\
        
        Naval Warfare &	This topic is about naval engagements and operations during various wars, focusing on the activities, missions, and modifications of naval ships and fleets. It includes minelaying, escort duties, battles, and ship modifications across different time periods and conflicts. \\

        Naval Armament	& This topic is about the various armaments and defensive features used in naval ships across different eras, including guns, torpedoes, and armor configurations. \\
        
        Naval Propulsion & This topic is about the propulsion systems and specifications of various naval ships, detailing their power sources, engine types, speeds, fuel capacities, and ranges. It encompasses the technological aspects of how these ships were designed to achieve their operational capabilities and endurance at sea. \\
        
    \bottomrule
    \end{tabular}
    \end{center}
    \label{tab:topic_def_pub}
\end{table}
\clearpage

\subsection{Business Dataset Examples}~\label{appendix:biz_example}

\begin{table}[h]
    \caption{Examples of business datasets and their corresponding topics generated by \textsc{SeLATM}.}
    \begin{center}
    \renewcommand{\arraystretch}{1.5}
    \footnotesize{
        \centering{\setlength\tabcolsep{2.0pt}}
    }

    \begin{tabular}{>{\centering\arraybackslash}m{2.0cm}>{\arraybackslash}m{8.0cm}>{\arraybackslash}m{4.0cm}}
        \toprule
        \textbf{Dataset} & \makecell[c]{\textbf{Input Text}} & \makecell[c]{\textbf{Generated Topic}} \\ \midrule

        CCC & VIRTUAL\_AGENT : Hi [Customer\_Name]. How can I help you? \newline CUSTOMER : Hi, how to avoid overdrawn fee \newline VIRTUAL\_AGENT : Which account would you like to get info on fees for? We want to make sure you're not surprised by a fee you didn't expect. \newline VIRTUAL\_AGENT : Here's where you can learn about the fees associated with your account, plus tips on avoiding some of them.        	& \{`Chatbot Greetings': 0.25, `Overdraft Fees': 0.25, `Fee Notifications,: 0.25, `Fee Information': 0.25\} \\

        BCR & I kept getting the same response to an issue different than the one I'm asking about & \{`Repetitive Responses': 1.0\} \\ 

        CI & Customer noticed a \$109 charge on their Apple bill pay for an in-person transaction in California, which they did not recognize. & \{`Apple Pay Issues': 1.0\} \\
        
    \bottomrule
    \end{tabular}
    \end{center}
    \label{tab:example_biz}
\end{table}

\begin{table}[h]
    \caption{The descriptions of topics presented in Table~\ref{tab:example_biz}.}
    \begin{center}
    \renewcommand{\arraystretch}{1.5}
    \footnotesize{
        \centering{\setlength\tabcolsep{2.0pt}}
    }

    \begin{tabular}{>{\arraybackslash}m{3.5cm}>{\arraybackslash}m{12.0cm}}
        \toprule
        \makecell[c]{\textbf{Topic}} & \makecell[c]{\textbf{Generated Topic Description}} \\ \midrule
        Chatbot Greetings & This topic is about the variety of introductory messages used by the bank's virtual agent to greet customers and initiate conversations, highlighting the personal touch in customer service interactions. \\

        Overdraft Fees & This topic is about customers' concerns and inquiries regarding overdraft fees, including requests for fee reversals, waivers, protection, and clarifications on charges. Customers often seek assistance to avoid or dispute these fees and express dissatisfaction with unexpected charges. \\

        Fee Notifications & This topic is about the bank's virtual agent ensuring that customers are informed about fees that might appear on their accounts, preventing any unexpected charges. It reflects the bank's proactive approach in maintaining transparency with its clients regarding potential fees. \\

        Fee Information &	This topic is about providing information on bank account fees and offering advice on how to avoid them. It focuses on educating customers about the costs associated with their accounts and helping them manage or reduce potential charges. \\

        Repetitive Responses & This topic is about users experiencing frustration with the chatbot providing repetitive and unhelpful responses, requiring them to rephrase questions multiple times without improvement in the outcome. \\
        
        Apple Pay Issues& This topic is about various difficulties customers face with Apple Pay, including unauthorized charges, declined transactions, and problems with account activation or verification. \\
    \bottomrule
    \end{tabular}
    \end{center}
    \label{tab:topic_def_biz}
\end{table}




\end{document}